\documentclass{article}
\usepackage{spconf,amsmath,graphicx}
\usepackage{subcaption}
\usepackage{float}
\ninept

\title{Transductive Kernels for Gaussian Processes on Graphs}
\name{Yin-Cong Zhi$^\star$ \qquad Felix L. Opolka$^{\dagger\mathsection}$ \qquad Yin Cheng Ng$^\ddagger$ \qquad Pietro Li\`o$^\dagger$ \qquad Xiaowen Dong$^\star$}
\address{$^{\star}$University of Oxford \qquad $^{\dagger}$University of Cambridge \qquad $^{\ddagger}$Man AHL \qquad $^{\mathsection}$Invenia Labs}

\newcommand{\res}[2]{$#1$ \small $\pm #2$}
\begin{document}
%
\maketitle
\begin{abstract}
Kernels on graphs have had limited options for node-level problems. To address this, we present a novel, generalized kernel for graphs with node feature data for semi-supervised learning. The kernel is derived from a regularization framework by treating the graph and feature data as two Hilbert spaces. We also show how numerous kernel-based models on graphs are instances of our design. A kernel defined this way has transductive properties, and this leads to improved ability to learn on fewer training points, as well as better handling of highly non-Euclidean data. We demonstrate these advantages using synthetic data where the distribution of the whole graph can inform the pattern of the labels. Finally, by utilizing a flexible polynomial of the graph Laplacian within the kernel, the model also performed effectively in semi-supervised classification on graphs of various levels of homophily.
\end{abstract}
\section{Introduction}
\label{sec:intro}

Graphs have become effective tools to represent more complex data by capturing the connectivity between points, as a result learning on such data has led to many performance improvements. While machine learning models can learn on just feature data, the graph structure can often enrich the model further. With the rise of graph signal processing \cite{shuman2012emerging, stankovic2019introduction}, we now have a rich library of tools to build models for graph structured data.

Gaussian processes (GP) are flexible tools for Bayesian modelling, and recent studies into building GPs on graphs have proved to be competitive against state-of-the-art graph neural network (GNN) models, giving rise to various kernel-based approaches for semi-supervised classification in \cite{Ng18,opolka2022adaptive,borovitskiy2021matern, liu2020uncertainty}, as well as other applications shown in \cite{Venkitaraman20, Zhi20, Opolka20, Walker19, Li20}. In this work, we focus on semi-supervised problems where the set of nodes are only partially labelled, and we aim to predict the unlabelled nodes. Graphs data of this form often comes with node attributes, which we will use as feature data, and in designing a kernel for a graph GP, the challenge is in embedding the graph connectivity into the kernel along with the feature data.

The most representative approaches to building GP on graphs for semi-supervised classification are \cite{Ng18,opolka2022adaptive}, where the authors made use of a matrix transformation on a graph-less base kernel on the node feature data. This could be a limiting factor for the model as the graph information comes into the model through a linear transformation. Furthermore, in this work we found if the base kernel is not particularly suitable, adding the graph elements will not be effective, and these models do not have a way to adjust the influence of the graph against the feature data kernel.

To address these issues, we use the principles of kernel design through regularization to derive a kernel with the ability to naturally handle the graph and feature data. In this approach, kernels are obtained by choosing regularization functions and finding a reproducing kernel Hilbert space (RKHS) \cite{smola1998connection}, typical continuous kernels such as RBF and Mat{\'e}rn kernels can all be derived in this manner.

The regularization approach has been translated to the graph domain in \cite{Smola03}, but currently this kernel depends on the graph only, and cannot incorporate node data. However, this method can be extended to graphs with node feature data, and in this paper we show how this can be achieved. We start by introducing the regularization approach to derive kernels for feature data and for graphs separately. We then present our approach by combining the two regularizers to obtain a kernel for graphs with node data, leading to our proposed model. The resulting model has transductive properties, meaning it trains on all nodes data in the graph, but only the training node labels. In addition, our kernel provides a clear way to control the influence of the graph compared to the feature data kernel. We then show that our setup is general, and that many graph learning models are actually instances of our design. We demonstrate the advantages of our model on synthetic data that are highly non-Euclidean and the training set is small. Lastly, we test on various real world graph-data in semi-supervised classification, comparing against various graph GP and popular GNN models.


\section{Background}
\label{background}

\begin{table*}[t]
    \tabcolsep=0.3cm
    \centering
    \begin{tabular}{llcc}
        \hline 
        \textbf{Model} & \textbf{Name} & $r_1(\Delta)$ & $r_2(\mathbf{L})$\\
        \hline
        Label Propagation & Label Propagation \cite{zhou2004regularization} & 0 & $\frac{1}{1-\alpha}(\mathbf{I} + \alpha\mathbf{L})$ \\
        Kernels on graphs & Regularized Laplacian \cite{Smola03} & 0 & $(\mathbf{I} + \sigma^2 \mathbf{L})$ \\
        & Diffusion \cite{Smola03} & 0 & $\exp\{\frac{\sigma^2}{2} \mathbf{L}\}$ \\
        & $p$-step random walk \cite{Smola03} & 0 & $(\alpha \mathbf{I} - \mathbf{L})^{-p}$ \\
        & Cosine \cite{Smola03} & 0 & $(\cos(\mathbf{L}\pi/4))^{-1}$ \\
        GP kernels & Mat{\'e}rn kernel on graphs \cite{borovitskiy2020matern} & 0 & $(\frac{2\nu}{\kappa^2} - \mathbf{L})^{\nu/2 + d/4}$\\
        & Laplacian kernel \cite{rasmussen2005gp} & $1 + ||\Delta||^2$ & 0 \\
         & Gaussian kernel \cite{rasmussen2005gp} & $e^{\frac{\sigma^2}{2} ||\Delta||^2}$ & 0 \\
        & Mat{\'e}rn kernel on manifolds \cite{borovitskiy2020matern} & $(\frac{2\nu}{\kappa^2} - \Delta)^{\nu/2 + d/4}$ & 0\\
        Graph GP & GGP \cite{Ng18} & $(\mathbf{P}^\top)^{-1} r(\Delta) \mathbf{P}^{-1}$ & 0 \\
        Wavelet Graph GP & WGGP \cite{opolka2022adaptive} & $(\mathbf{W}^\top)^{-1} r(\Delta) \mathbf{W}^{-1}$ & 0 \\
        Transductive kernel (ours) & TGGP (ours) & $(\frac{2\nu}{\kappa^2} - \Delta)^{\nu/2 + d/4}$ & $\mathbf{U} [\text{softplus}(\sum_i(\beta_i\mathbf{\Lambda}^i))] \mathbf{U}^\top$ \\
        \hline
    \end{tabular}
    \caption{Various types of graph learning models that can be considered instances of the transductive GP. For GGP and WGGP, the matrices are $\mathbf{P} = (\mathbf{I} + \mathbf{D})^{-1}(\mathbf{I} + \mathbf{A})$ and $\mathbf{W} = h(\mathbf{L}) + g_1(\mathbf{L}) + g_2(\mathbf{L})$ for low-pass $h$ and band-pass $g_1$ and $g_2$ defined in \cite{opolka2022adaptive}. $r(\Delta)$ is not defined for these two models due to the kernel used being a dot product kernel, which does not have a corresponding regularization function.}\label{related_models}
\end{table*}

\subsection{The Graph Laplacian}

A graph dataset $\mathcal{G}$ can be defined by $(\mathcal{V},\mathbf{A},(\mathbf{X},\mathbf{y}))$, where $\mathcal{V} = \{v_1,\dots,v_N\}$ is the set of vertices each with associated node attributes $\mathbf{x}_i \in \mathbf{X}$ and label $y_i \in \mathbf{y}$, and $\mathbf{A}$ is the symmetric adjacency matrix, we will refer $\mathbf{x}_i$ as the feature data of the node.

In the literature of graph signal processing \cite{shuman2012emerging, stankovic2019introduction}, the notion of smoothness and spectral analysis have been used to analyze behaviours of signals on graphs; in particular, they are governed by the graph Laplacian, where the normalized version is defined as
\begin{align}
    \mathbf{L} = \mathbf{D}^{-\frac{1}{2}} [\mathbf{D} - \mathbf{A}] \mathbf{D}^{-\frac{1}{2}}
\end{align}
for diagonal degree matrix $\mathbf{D}$.

If we interpret graphs as Hilbert spaces, then the Laplacian provides a measure of smoothness within the space. In particular, in \cite{Smola03} it has been shown that the graph Laplacian is the discretized version of the Laplace operator $\Delta = \sum_j\frac{\partial^2 }{\partial x_j^2}$ where $x_j$ is the $j$th element of a feature vector $\mathbf{x}$. Thus, many functions derived from $\Delta$ can be translated to the graph domain by replacing the Laplace operator with the graph Laplacian (e.g. \cite{Smola03, borovitskiy2021matern}).

\subsection{Kernels via Regularization}

Many machine learning models can be formulated as a minimization problem of a loss function, and often regularizers can be added to the objective to enforce certain structures on the model (for instance to avoid over-fitting). By examining the regularizer, kernel functions can then be derived by assuming that solution comes from a RKHS. In the continuous domain, as described by \cite{smola1998connection}, we start with the following regularization problem
\begin{align}
\arg\min_{f} \quad \text{loss}(f,\mathbf{y}) + \Omega(||f||^2) \label{loss_function}
\end{align}
between model $f$ as a function of $\mathbf{x}_i$, and labels $\mathbf{y}$. We consider the regularizing function that can be represented by an inner product
\begin{align}
\Omega(||f||^2) = \langle f, f \rangle_\mathcal{H} = \langle Pf, Pf \rangle = \langle f, r(\Delta) f \rangle. \label{norm}
\end{align}
We write the norm in this form where $P$ is a collection of smoothness measures on $f$, for example $P = 1, \nabla = [\frac{\partial}{\partial x_1}, \dots, \frac{\partial}{\partial x_m}]^\top$ or $(1, \nabla)^\top$. Due to $\nabla . \nabla = \Delta = \sum_i \frac{\partial^2}{\partial x_i^2}$ and $\langle Pf, Pf \rangle = \langle f, P^\top.Pf \rangle$, we can combine $P^\top.P$ into a function of $\Delta$, and the norm can be characterized by the function $r(\Delta)$. 

The inner product then specifies a Hilbert space $\mathcal{H}$, and through Representer Theorem, we know that if there exists a kernel such that $\mathcal{H}$ is a RKHS, then there exist a solution to (\ref{loss_function}) that is a linear combination of the kernel $f^*(\mathbf{x}') = \sum_i w_i k(\mathbf{x}_i,\mathbf{x}')$. We will refer readers to \cite{hofmann2008kernel} for more details on kernel methods.

By deriving a kernel $k$ that makes (\ref{norm}) a RKHS, we require the reproducing property on the norm as follows
\begin{align}
    \langle f(\mathbf{x}), k(\mathbf{x}, \mathbf{x}') \rangle_\mathcal{H} = f(\mathbf{x}'), \; \forall \mathbf{x}'.
\end{align}
This can be written as
\begin{align}
    \langle f(\mathbf{x}), r(\Delta) k(\mathbf{x}, \mathbf{x}') \rangle &= \int f(\mathbf{x}) r(\Delta)k(\mathbf{x}, \mathbf{x}') \; d\mathbf{x} = f(\mathbf{x}').
\end{align}
If we look at the terms inside the integral, the above problem is equivalent to finding $k(\mathbf{x}, \cdot)$ such that $k(\mathbf{x}, \mathbf{x}') r(\Delta) = \delta(\mathbf{x}-\mathbf{x}')$
where $\delta$ is the Dirac delta function. Finding $k$ boils down to finding the Green's function and the solution is generally written as $k(\mathbf{x}, \mathbf{x}') = r^{-1}(\Delta)_{(\mathbf{x} - \mathbf{x}')}$, or in matrix format
\begin{align}
\mathbf{K} = r^{-1}(\Delta).
\end{align}
Examples of this derivation can be found in \cite{smola1998connection}, note above equation is now a function of $\mathbf{x}$.

\begin{figure*}[h]
\centering
\begin{subfigure}[b]{0.31\textwidth}
    \centering
    \includegraphics[width=1.\linewidth, height = 3.6cm]{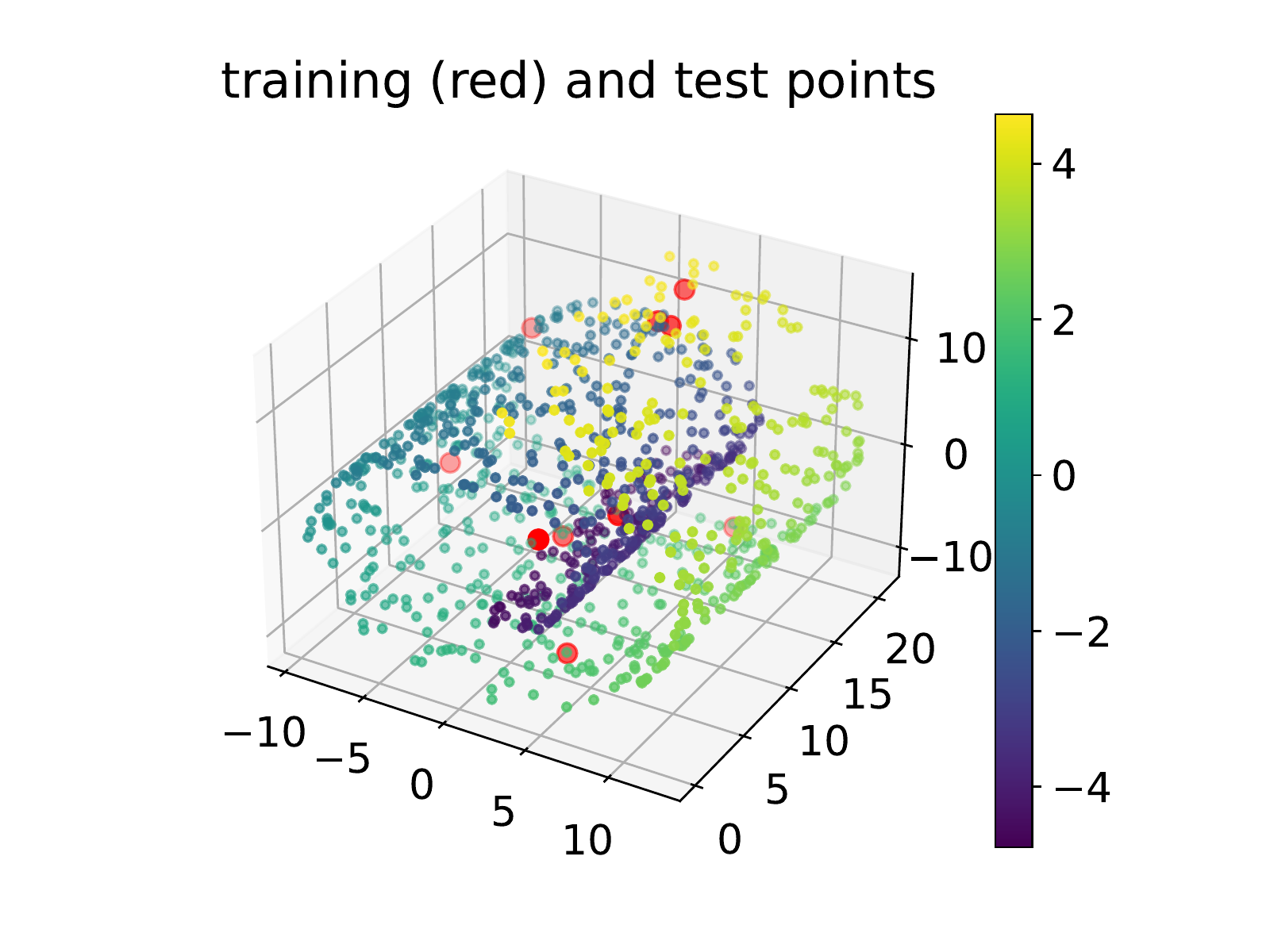}
    \caption{Ground truth and training points in red}\label{ground_truth}
\end{subfigure}
\begin{subfigure}[b]{0.31\textwidth}
    \centering
    \includegraphics[width=1.\linewidth, height = 3.6cm]{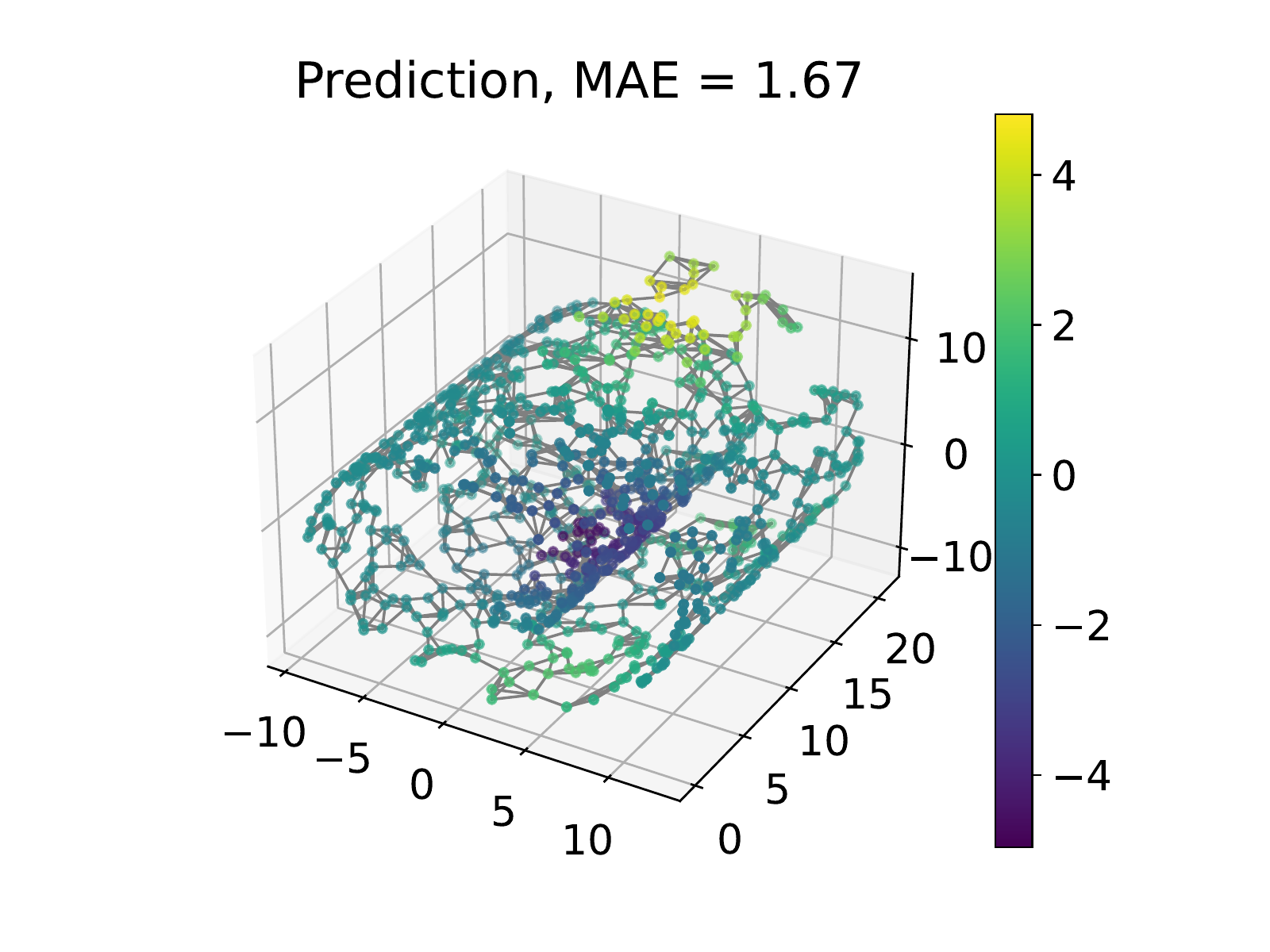}
    \caption{GP with RBF}\label{gp_only}
\end{subfigure}
\begin{subfigure}[b]{0.31\textwidth}
    \centering
    \includegraphics[width=1.\linewidth, height = 3.6cm]{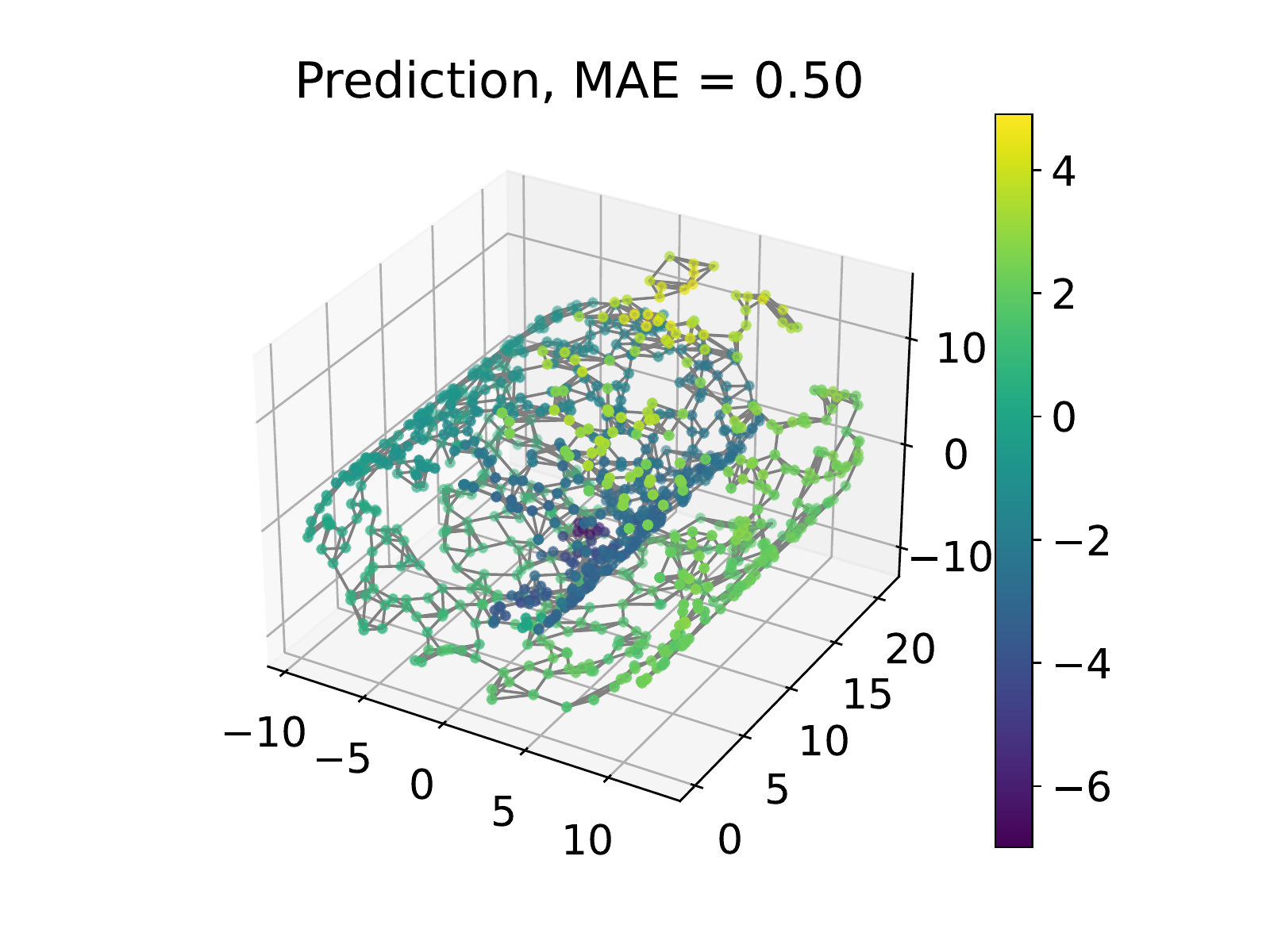}
    \caption{Graph only $(\mathbf{I} + \alpha \mathbf{L})^{-1}$}\label{graph_only}
\end{subfigure}
\begin{subfigure}[b]{0.31\textwidth}
    \centering
    \includegraphics[width=1.\linewidth, height = 3.6cm]{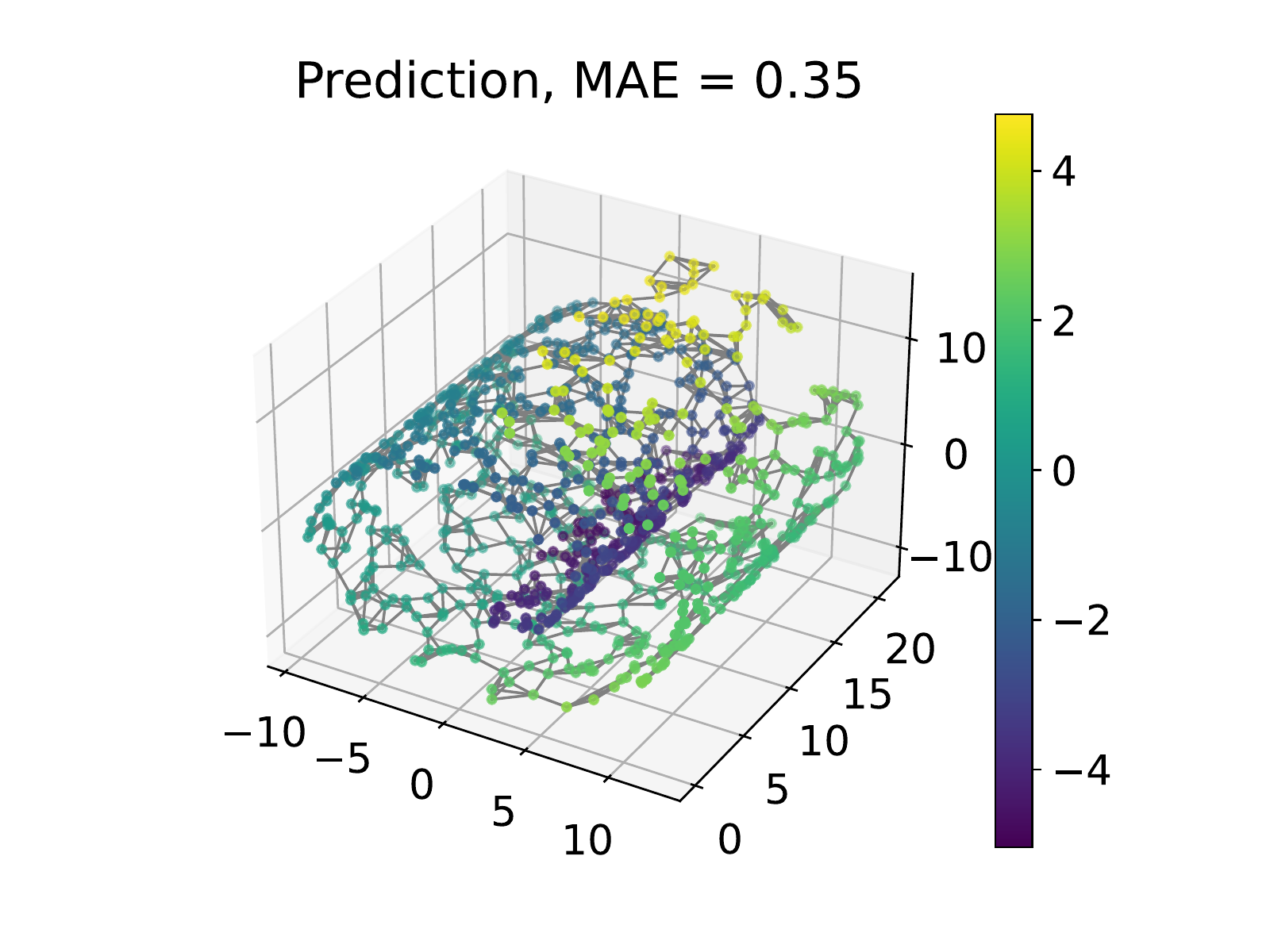}
    \caption{TGGP (ours)}
\end{subfigure}
\begin{subfigure}[b]{0.31\textwidth}
    \centering
    \includegraphics[width=1.\linewidth, height = 3.6cm]{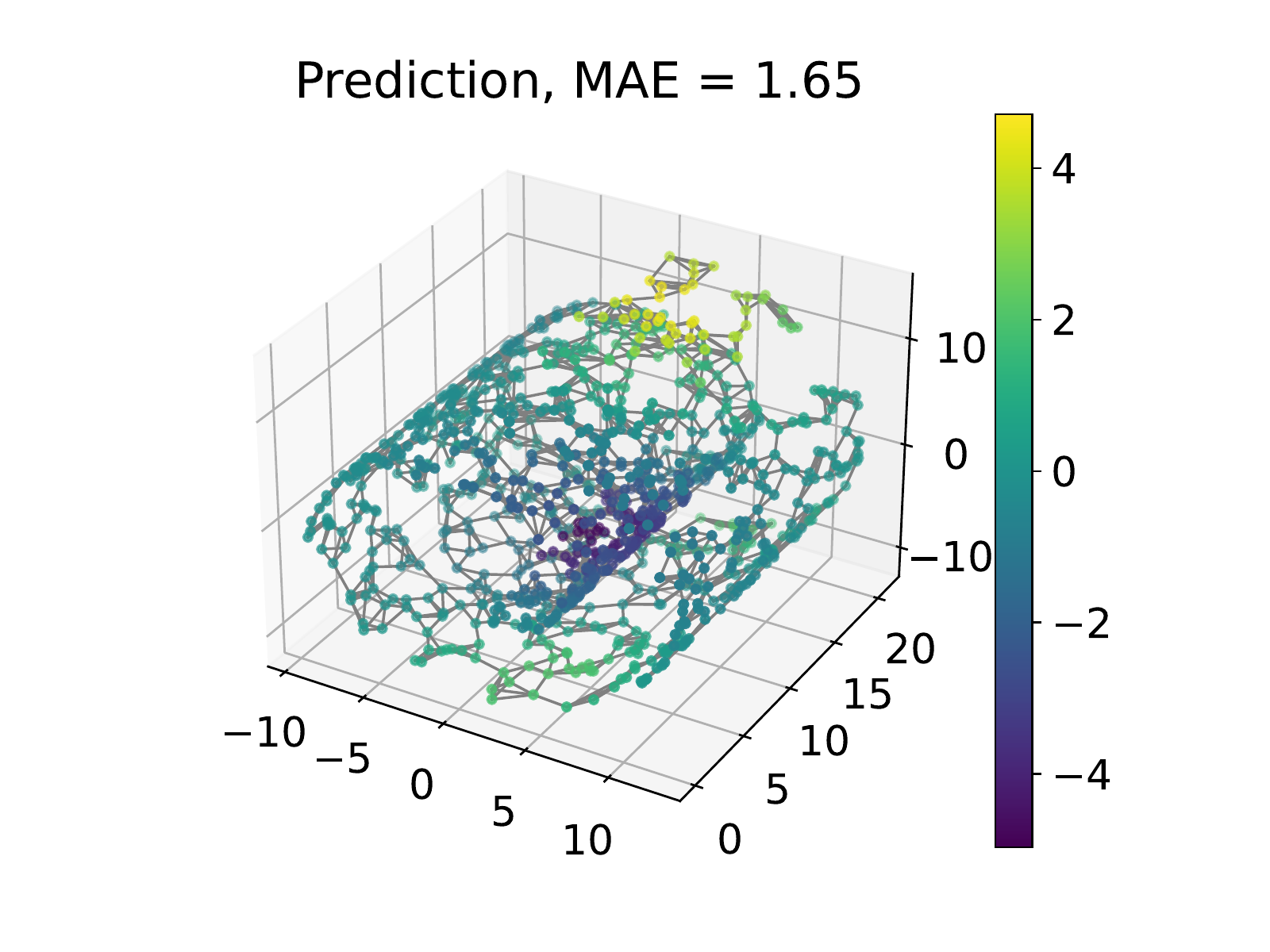}
    \caption{GGP \cite{Ng18}}
\end{subfigure}
\begin{subfigure}[b]{0.31\textwidth}
    \centering
    \includegraphics[width=1.\linewidth, height = 3.6cm]{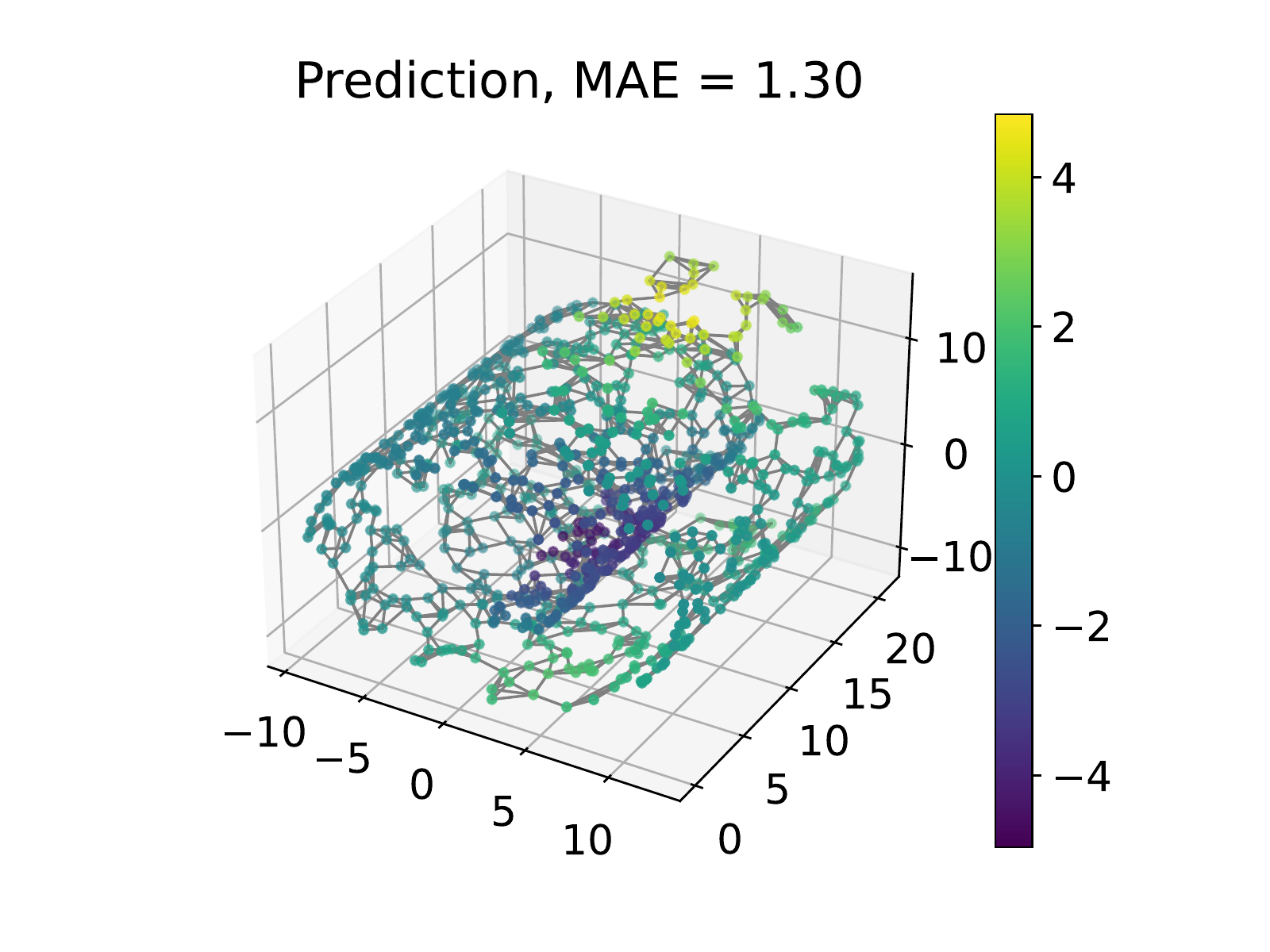}
    \caption{WGGP \cite{opolka2022adaptive}}
\end{subfigure}
\caption{Swiss roll regression training on 10 points. Each plot shows the prediction from a GP model and the MAE is computed against ground truth in (a). Model (d) best resembled the ground truth and produced the lowest MAE. Results from repeated runs are presented in Fig. \ref{bound}}\label{synthetic}
\end{figure*}

\subsection{Kernels and Regularization on Graphs}

We now detail how the regularization approach to deriving kernels has been translated to the graph domain in \cite{Smola03} by exploring the equivalence between the Laplace operator and the graph Laplacian.

Let $\mathbf{f}$ be a function or signal on the graph, we again start from the regularization problem of (\ref{loss_function}), but instead of $\Delta$ the regularizer is now replaced by the graph Laplacian
\begin{align}
    \Omega(||\mathbf{f}||^2) = \langle \mathbf{f}, r(\mathbf{L}) \mathbf{f} \rangle = \mathbf{f}^\top r(\mathbf{L}) \mathbf{f}
\end{align}
where the discrete nature of graphs means the inner product becomes matrix multiplication instead of integration. Next, we again require the reproducing property, we do this by looking at the $i$th element
\begin{align}
    [\mathbf{f}^\top r(\mathbf{L}) \mathbf{K}]_i = \mathbf{f}_i \; \forall i,
\end{align}
and we can compute the kernel on a graph as follows
\begin{align}
    \mathbf{f}^\top r(\mathbf{L}) \mathbf{K} = \mathbf{f} \implies \mathbf{K} = r^{-1}(\mathbf{L}).
\end{align}
The kernel derived here is a measure of similarity between nodes using solely the connectivities of the graph, and not on any node feature data. In the next section, we introduce how to derive a kernel for nodes that is also dependent on any additional feature data.

\section{Transductive Kernels for Graphs with Feature Data}
\label{model}

Most graphs have feature data $\mathbf{x}_i$ associated with each node $v_i$. Thus, there is the need to find a kernel of the form $k(\mathbf{x}_i, \mathbf{x}_j |\mathcal{G})$. We do this by noting that $\mathbf{x}_i$ belongs to a continuous space, while the nodes fall on the graph space. Thus, if we are to take into account both spaces of information, we want to use two regularizers
\begin{align}
    \Omega(||\mathbf{f}||^2) = \langle \mathbf{f}, [r_1(\Delta) + r_2(\mathbf{L})] \mathbf{f} \rangle.
\end{align}
Like the previous section, by the reproducing property, the kernel is
\begin{align}
    \mathbf{K} = [r_1(\Delta) + r_2(\mathbf{L})]^{-1}. \label{rkernel}
\end{align}
Although we can simply take the $(i,j)$th element of the inverse to obtain $k(\mathbf{x}_i, \mathbf{x}_j | \mathcal{G})$, we can write this in a more explicit form by using the Woodbury formula \cite{petersen2008matrix}, first setting $\mathbf{K}_1 = r^{-1}_1(\Delta)$,
\begin{align}
    [\mathbf{K}_1^{-1} + r_2(\mathbf{L})]^{-1} = \mathbf{K}_1 - \mathbf{K}_1[\mathbf{I} + r_2(\mathbf{L})\mathbf{K}_1]^{-1} r_2(\mathbf{L}) \mathbf{K}_1
\end{align}
from which we can see that elementwise
\begin{align}
k(\mathbf{x}_1, \mathbf{x}_2) = &k_1(\mathbf{x}_1, \mathbf{x}_2) - \nonumber\\
&k_1(\mathbf{x}_1, \mathbf{X})^\top [\mathbf{I} + r_2(\mathbf{L})\mathbf{K}_1]^{-1} r_2(\mathbf{L}) k_1 (\mathbf{X}, \mathbf{x}_2) \label{transductive}
\end{align}
where $\mathbf{X}$ is the matrix of feature data on all the nodes.

The notable feature of (\ref{transductive}) is that the kernel between any two points depends on the full graph Laplacian and the feature data on every node. This gives the kernel the transductive property, as during training the model has access to both the training and test nodes data (and any potential unlabelled nodes), but only the training labels. This is similar to the resulting kernel of \cite{kim2014transductive}, but that approach is in modifying the Hilbert space by weight construction between points, and not based on separate graph information.

The transductive nature of the kernel is due to the inversion in (\ref{transductive}), this means to evaluate the prior there is a complexity of $\mathcal{O}(N^3)$ for size of the graph $N$. This is currently a limitation and therefore this kernel can only handle graphs of reasonable sizes.

\begin{table*}[h]
    \tabcolsep=0.2cm
    \centering
    \begin{tabular}{lcccccccc}
        \hline
        \textbf{Method} & \textbf{Texas} & \textbf{Cornell} & \textbf{Wisconsin} & \textbf{Chameleon} & \textbf{Cora} & \textbf{Citeseer} & \textbf{Squirrel} & \textbf{Actor}\\
        \textbf{\# Nodes} & 183 & 183 & 251 & 2,277 & 2,708 &  3,327 & 5,201 & 7,600\\
        \textbf{Homophily Ratio} & 0.11 & 0.30 & 0.21 & 0.23 & 0.81 & 0.74 & 0.22 & 0.22\\
        \hline
        \textbf{GCN} \cite{kipf2017gcn} & \res{59.5}{5.3} & \res{57.0}{4.7} & \res{59.8}{7.0} & \res{59.8}{2.6} & \res{80.5}{0.8} & \res{68.1}{1.3} & \res{36.9}{1.3} & \res{30.3}{0.8}\\
        \textbf{GAT} \cite{velickovic2018gat} & \res{58.4}{4.5} & \res{58.9}{3.3} & \res{55.3}{8.7} & \res{54.7}{2.0} & \res{\textbf{82.6}}{\textbf{0.7}} & \res{\textbf{72.2}}{\textbf{0.9}} & \res{30.6}{2.1} & \res{26.3}{1.7}\\
        \textbf{ChebNet} \cite{defferrard2016chebnet} & \res{77.3}{4.1} & \res{\textbf{74.3}}{\textbf{7.5}} & \res{79.4}{4.5} & \res{55.2}{2.8} & \res{78.0}{1.2} & \res{70.1}{0.8} & \res{43.9}{1.6} & \res{34.1}{1.1}\\
        \textbf{LP} \cite{zhou2004regularization} & 37.8 & 21.6 & 23.5 & 44.5 & 71.3 & 49.9 & 32.7 & 22.4\\
        \textbf{GP} \cite{Ng18} & 78.4 & 73.0 & 78.4 & 46.1 & 60.8 & 54.7 & 34.4 & \textbf{34.9}\\
        \textbf{GGP} \cite{Ng18} & 78.4 & 62.1 & 60.8 & \textbf{73.5} & 80.9 & 69.7 & \textbf{64.8} & 26.3\\
        \textbf{ChebGP} \cite{opolka2022adaptive} & \textbf{81.1} & 64.9 & \textbf{82.4} & \textbf{69.1} & 79.7 & 66.5 & 28.8 & 31.8 \\
        \textbf{WGGP} \cite{opolka2022adaptive} & 78.4 & 67.6 & \textbf{84.3} & 64.5 & \textbf{84.7} & \textbf{70.8} & \textbf{58.3} & 32.6\\
        \textbf{TGGP} (ours) & \textbf{81.1} & \textbf{75.7} & \textbf{82.4} & 63.2 & 80.3 & 70.5 & 53.8 & \textbf{34.9}\\
        \hline
        \textbf{GGP-X} \cite{Ng18} & 78.4 & 56.8 & 60.8 & \textbf{77.6} & 84.7 & 75.6 & \textbf{71.9} & OOM \\
        \textbf{WGGP-X} \cite{opolka2022adaptive} & 81.1 & 75.7 & 84.3 & 65.6 & \textbf{87.5} & \textbf{76.8} & 61.3 & OOM \\
        \textbf{TGGP-X} (ours) & \textbf{86.5} & \textbf{81.1} & \textbf{86.3} & 63.4 & 83.8 & 76.7 & 54.2 & \textbf{36.9}\\
        \hline
    \end{tabular}
    \caption{Classification percentage accuracy on real world datasets. Entries with OOM are due to ``Out Of Memory" error and thus we were unable to run the experiment on the dataset. The two best performing models and the highest \textbf{-X} version
    are highlighted.} \label{realworld}
\end{table*}

\subsection{Flexible Modelling of Regularization on Graphs}

The regularizer $r_1(\Delta)$ is not always implicitly specified, in particular it does not exist for dot product kernels \cite{smola2000regularization}. Instead, we can directly choose the kernel function $\mathbf{K}_1$, and examples of $\mathbf{K}_1$ can be any typical kernel such as the RBF
\begin{align}
    \sigma_1^2 \exp\big\{-||\mathbf{x}_i - \mathbf{x}_j||^2/2l^2\big\}.
\end{align}
Meanwhile, examples of $r_2(\mathbf{L})$ are $r_2(\mathbf{L}) = \frac{1}{\sigma_2^2}(\mathbf{I} + \alpha\mathbf{L})$, or $\frac{1}{\sigma_2^2}\exp\{\alpha\mathbf{L}\}$ for  $\alpha>0$,
which along with the choice in \cite{kim2014transductive} are of low-pass nature \cite{Smola03}. A more flexible option is polynomials, and for this we first have $\mathbf{L} = \mathbf{U\Lambda U}^\top$, then the graph regularization term can be written as
\begin{align}
    r_2(\mathbf{L}) = \frac{1}{\sigma_2^2}\mathbf{U}r_2(\boldsymbol{\Lambda}) \mathbf{U}^\top
\end{align}
where $r_2$ can be defined as an element-wise function on the graph Laplacian eigenvalues $\lambda$, and $\sigma_2^2$ acts as the variance term of $r_2(\mathbf{L})$. The polynomial coefficients can be freely learnt during the optimization, but we require $r_2(\lambda) > 0$ for all $\lambda$ in order to avoid any 0 function eigenvalues which will make the kernel singular. We achieve this by passing the polynomial through a positive softplus function
\begin{align}
    r_2(\lambda) = \text{softplus}\bigg(\sum_{i=0}^d \beta_i \lambda^i\bigg)
\end{align}
where each $\beta_i$ is learnt freely, and $\text{softplus}(x) = \log(1 + e^x)$.

The variance terms $\sigma_1^2$ and $\sigma_2^2$ not only capture the variance in the data, but also act as weighting functions on the two terms in the kernel in (\ref{rkernel}). Thus, by observing the size of these two hyperparameters, we can also tell which element is more important between the node feature data and the graph structure.

\subsection{Relationships With Other Models}

Our proposed kernel in (\ref{rkernel}) is also a general framework for learning on graphs. Notably, when the two terms $r_1(\Delta)$ and $r_2(\mathbf{L})$ take certain forms, we recover well known models on graphs. We present several of these in Table \ref{related_models}. In particular, we note that all existing kernel-based models correspond to having only one of $r_1(\Delta)$ or $r_2(\mathbf{L})$, while our model is unique in that it combines the two.

\section{Experiments}
\label{experiments}

\begin{figure}[h]
\centering
    \centering
    \includegraphics[width=0.75\linewidth, height = 4.2cm]{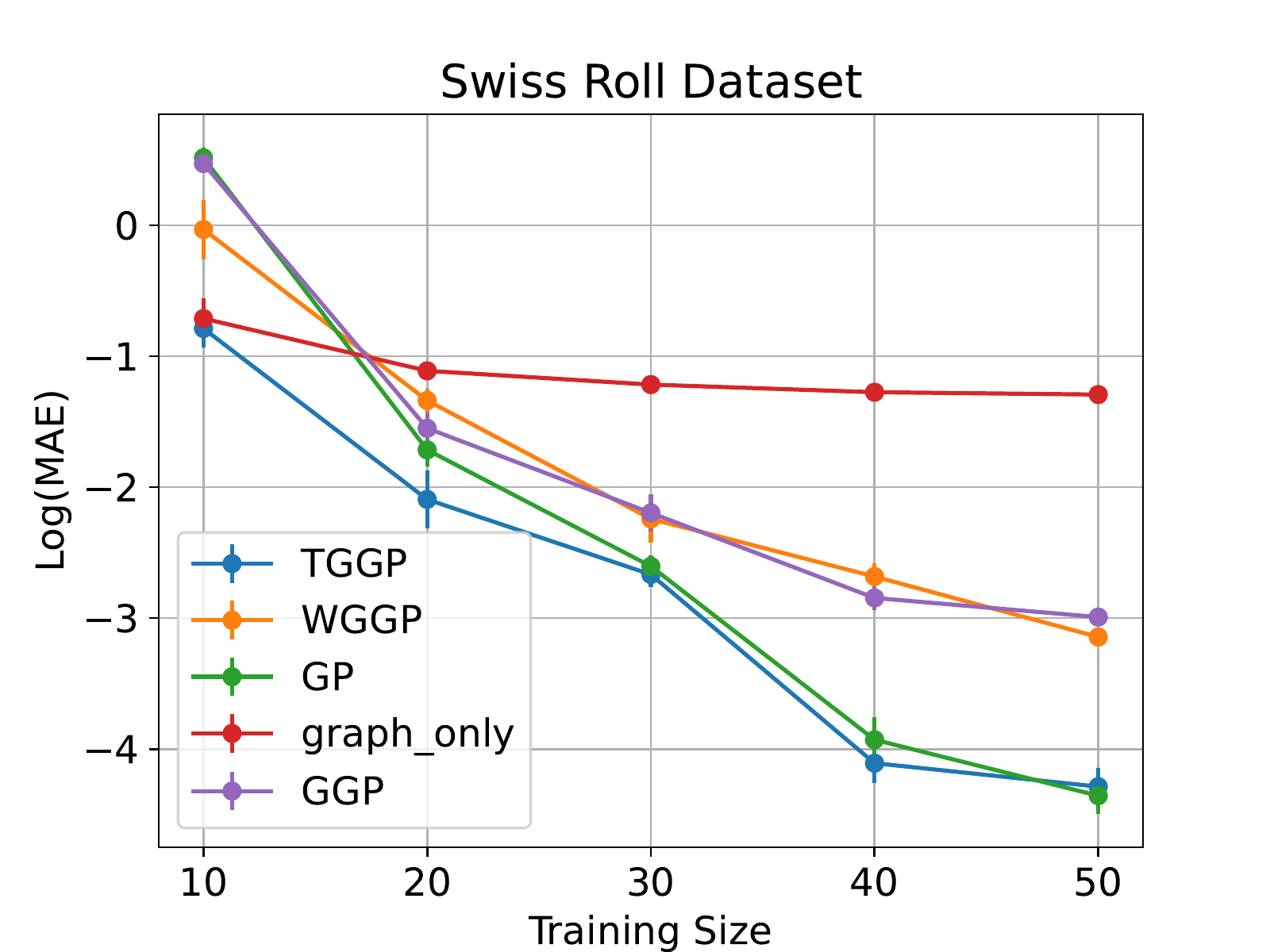}
    \caption{Mean MAE and standard error over 10 samples of Swiss roll synthetic data on different number of training nodes.}\label{bound}
\end{figure}

\subsection{Synthetic Results}
\label{synethetic}

We first experiment on the Swiss-Roll dataset, where its non-Euclidean nature gives the need for using a graph. We sample 1000 points from the roll-like surface on which we construct a $k$-nearest-neighbour graph, with $k = 4$ so that it is small while ensuring the graph is fully connected. Each point has a continuous label that is increasing as the roll moves from the inner-most end to the outer-most of the surface. The nodes have coordinates as attributes which we will use as feature data.

The transductive nature of the kernel heavily exploits the graph's ability to capture the roll-like structure, allowing the distribution of the test points to have a big effect on the prediction. As a result the model picks up more information than simpler models, and this advantage is most apparent on fewer training data when other models struggle to learn. To demonstrate this, we randomly use only 10 labels as training, and predict the labels on the remaining 990 points.

We compare against kernels using just the feature data (standard GP), and graph only \cite{Smola03} as sanity checks. We also test against GGP \cite{Ng18} and WGGP \cite{opolka2022adaptive}, as they are both kernel based models, but with a different setup to our proposed model. We plot the predictions on the graph, as well as the MAE against the test node labels in Fig. \ref{synthetic}.

Using the graph only kernel in Fig. \ref{graph_only} already improved on the MAE of using the \textbf{GP} alone in Fig. \ref{gp_only}, indicating the graph is more informative than the feature data, while our \textbf{TGGP} combines the two and was the best in terms of MAE and reproducing the pattern of the ground truth. Current graph GP models \textbf{GGP} and \textbf{WGGP} rely on the base kernel that is in \textbf{GP}, and therefore do not perform as well.

We further provide the log MAEs over multiple realizations of the Swiss-roll in Fig. \ref{bound}. As the training points increase, the \textbf{GP} starts to show its predictive power indicating the feature data is now more informative. Only \textbf{TGGP} continues to perform as well as \textbf{GP} as it has the ability to weigh the graph and feature data within the kernel.

\subsection{Real World Experiments}
\label{real}

We lastly perform semi-supervised classification on a number of real-world graph datasets. We take $\mathbf{A}' = \frac{1}{2}(\mathbf{A} + \mathbf{A}^\top)$ as the adjacency matrix if the graph is directed to ensure eigenvalues are real. The $r_2$ function is chosen as a degree 4 polynomial, and a Mat{\'e}rn12 is used for $\mathbf{K}_1$. We use the public split of training, validation, and testing sets, as generally the training set is smaller so this will showcase our model's ability to learn on fewer data. Much like the GP models of \cite{Ng18, opolka2022adaptive}, we also include the \textbf{-X} version where we include the validation labels in the training. As we are in a classification setting where uncertainty measure is less useful, we simply make use of a multi-output regression GP to predict continuous values for each class, then taking the argmax to obtain the prediction class. The classification accuracies are shown in Table \ref{realworld}, where we compared mainly against graph GP models as well as popular GNN models. \textbf{TGGP} was the best performing model in a number of datasets while still being amongst the top in others. \textbf{TGGP-X} also performed similarly compared to other \textbf{-X} models.

A way of measuring the smoothness of the data is in the homophily ratio, and graph-based models are known to perform well when the ratio is high such as Cora and Citeseer, on which \textbf{TGGP} performed competitively. The other datasets exhibited relatively low ratio, and it was clear that none of the baseline models performed consistently well on these datasets. \textbf{TGGP}, on the other hand, made use of the flexible polynomial to pick-up non-smooth elements in the data. As a result it was the only graph GP model that consistently improved on \textbf{GP} which does not make use of the graph in any way.

\section{Conclusion}

We have proposed an extensive definition of kernels on graphs that can take into account graph information and node feature data for semi-supervised learning. Using the graph inherently makes the kernel transductive, allowing the model to learn on the test information and capture the distribution of the full dataset. Our kernel design is generic, and we have showed that numerous kernels and Laplacian-based models can be considered an instance of our design. Our model showed superior ability to learn on synthetic data where we emphasized on small number of training points and a highly non-Euclidean nature. On real world graphs, we also showed competitive classification accuracies against various graph GP and GNN models.

\bibliographystyle{IEEEbib}
\bibliography{ref.bib}

\begin{thebibliography}{10}

\bibitem{shuman2012emerging}
David~I Shuman, Sunil~K Narang, Pascal Frossard, Antonio Ortega, and Pierre
  Vandergheynst,
\newblock ``The emerging field of signal processing on graphs: Extending
  high-dimensional data analysis to networks and other irregular domains,''
\newblock {\em IEEE signal processing magazine}, vol. 30, no. 3, pp. 83--98,
  2013.

\bibitem{stankovic2019introduction}
Ljubi{\v{s}}a Stankovi{\'c}, Milo{\v{s}} Dakovi{\'c}, and Ervin Sejdi{\'c},
\newblock ``Introduction to graph signal processing,''
\newblock in {\em Vertex-Frequency Analysis of Graph Signals}, pp. 3--108.
  Springer, 2019.

\bibitem{Ng18}
Yin~Cheng Ng, Nicol{\`o} Colombo, and Ricardo Silva,
\newblock ``Bayesian semi-supervised learning with graph gaussian processes,''
\newblock {\em Advances in Neural Information Processing Systems}, vol. 31,
  2018.

\bibitem{opolka2022adaptive}
Felix Opolka, Yin-Cong Zhi, Pietro Li{\`o}, and Xiaowen Dong,
\newblock ``Adaptive gaussian processes on graphs via spectral graph
  wavelets,''
\newblock in {\em International Conference on Artificial Intelligence and
  Statistics}. PMLR, 2022, pp. 4818--4834.

\bibitem{borovitskiy2021matern}
Viacheslav Borovitskiy, Iskander Azangulov, Alexander Terenin, Peter Mostowsky,
  Marc Deisenroth, and Nicolas Durrande,
\newblock ``Mat{\'e}rn gaussian processes on graphs,''
\newblock in {\em International Conference on Artificial Intelligence and
  Statistics}. PMLR, 2021, pp. 2593--2601.

\bibitem{liu2020uncertainty}
Zhao-Yang Liu, Shao-Yuan Li, Songcan Chen, Yao Hu, and Sheng-Jun Huang,
\newblock ``Uncertainty aware graph gaussian process for semi-supervised
  learning,''
\newblock in {\em Proceedings of the AAAI Conference on Artificial
  Intelligence}, 2020, vol.~34, pp. 4957--4964.

\bibitem{Venkitaraman20}
Arun Venkitaraman, Saikat Chatterjee, and Peter Handel,
\newblock ``Gaussian processes over graphs,''
\newblock in {\em ICASSP 2020-2020 IEEE International Conference on Acoustics,
  Speech and Signal Processing (ICASSP)}. IEEE, 2020, pp. 5640--5644.

\bibitem{Zhi20}
Yin-Cong Zhi, Yin~Cheng Ng, and Xiaowen Dong,
\newblock ``Gaussian processes on graphs via spectral kernel learning,''
\newblock {\em arXiv preprint arXiv:2006.07361}, 2020.

\bibitem{Opolka20}
Felix~L Opolka and Pietro Li{\`o},
\newblock ``Graph convolutional gaussian processes for link prediction,''
\newblock {\em arXiv preprint arXiv:2002.04337}, 2020.

\bibitem{Walker19}
Ian Walker and Ben Glocker,
\newblock ``Graph convolutional gaussian processes,''
\newblock in {\em International Conference on Machine Learning}. PMLR, 2019,
  pp. 6495--6504.

\bibitem{Li20}
Naiqi Li, Wenjie Li, Jifeng Sun, Yinghua Gao, Yong Jiang, and Shu-Tao Xia,
\newblock ``Stochastic deep gaussian processes over graphs,''
\newblock {\em Advances in Neural Information Processing Systems}, vol. 33, pp.
  5875--5886, 2020.

\bibitem{smola1998connection}
Alex~J Smola, Bernhard Sch{\"o}lkopf, and Klaus-Robert M{\"u}ller,
\newblock ``The connection between regularization operators and support vector
  kernels,''
\newblock {\em Neural networks}, vol. 11, no. 4, pp. 637--649, 1998.

\bibitem{Smola03}
Alexander~J Smola and Risi Kondor,
\newblock ``Kernels and regularization on graphs,''
\newblock in {\em Learning theory and kernel machines}, pp. 144--158. Springer,
  2003.

\bibitem{zhou2004regularization}
Dengyong Zhou and Bernhard Sch{\"o}lkopf,
\newblock ``A regularization framework for learning from graph data,''
\newblock in {\em ICML 2004 Workshop on Statistical Relational Learning and Its
  Connections to Other Fields (SRL 2004)}, 2004, pp. 132--137.

\bibitem{borovitskiy2020matern}
Viacheslav Borovitskiy, Alexander Terenin, Peter Mostowsky, et~al.,
\newblock ``Mat{\'e}rn gaussian processes on riemannian manifolds,''
\newblock {\em Advances in Neural Information Processing Systems}, vol. 33, pp.
  12426--12437, 2020.

\bibitem{rasmussen2005gp}
Carl~Edward Rasmussen and Christopher K.~I. Williams,
\newblock {\em Gaussian Processes for Machine Learning (Adaptive Computation
  and Machine Learning)},
\newblock The MIT Press, 2005.

\bibitem{hofmann2008kernel}
Thomas Hofmann, Bernhard Sch{\"o}lkopf, and Alexander~J Smola,
\newblock ``Kernel methods in machine learning,''
\newblock {\em The annals of statistics}, vol. 36, no. 3, pp. 1171--1220, 2008.

\bibitem{petersen2008matrix}
Kaare~Brandt Petersen, Michael~Syskind Pedersen, et~al.,
\newblock ``The matrix cookbook,''
\newblock {\em Technical University of Denmark}, vol. 7, no. 15, pp. 510, 2008.

\bibitem{kim2014transductive}
Hyun-Chul Kim, Jaewook Lee, and Daewon Lee,
\newblock ``Transductive gaussian processes with applications to object pose
  estimation,''
\newblock {\em The Computer Journal}, vol. 57, no. 3, pp. 339--346, 2014.

\bibitem{kipf2017gcn}
Thomas~N. Kipf and Max Welling,
\newblock ``{Semi-Supervised Classification with Graph Convolutional
  Networks},''
\newblock in {\em International Conference on Learning Representations}, 2017.

\bibitem{velickovic2018gat}
Petar Veličković, Guillem Cucurull, Arantxa Casanova, Adriana Romero, Pietro
  Liò, and Yoshua Bengio,
\newblock ``Graph attention networks,''
\newblock in {\em International Conference on Learning Representations}, 2018.

\bibitem{defferrard2016chebnet}
Micha\"{e}l Defferrard, Xavier Bresson, and Pierre Vandergheynst,
\newblock ``Convolutional neural networks on graphs with fast localized
  spectral filtering,''
\newblock in {\em Advances in Neural Information Processing Systems 29}, 2016,
  pp. 3844--3852.

\bibitem{smola2000regularization}
Alex Smola, Zolt{\'a}n Ov{\'a}ri, and Robert~C Williamson,
\newblock ``Regularization with dot-product kernels,''
\newblock {\em Advances in neural information processing systems}, vol. 13,
  2000.

\end{thebibliography}

\end{document}